\documentclass{article}

\usepackage{arxiv}

\usepackage[utf8]{inputenc} 
\usepackage[T1]{fontenc}    
\usepackage{hyperref}       
\usepackage{url}            
\usepackage{booktabs}       
\usepackage{amsfonts}       
\usepackage{nicefrac}       
\usepackage{microtype}      
\usepackage{lipsum}		
\usepackage{graphicx}
\usepackage{natbib}
\usepackage{doi}

\usepackage{amsmath}
\usepackage{subfig}
\usepackage{setspace}
\usepackage[linesnumbered,ruled,vlined]{algorithm2e}
\usepackage{multirow}
\usepackage{hhline}
\usepackage{graphicx}
\def \hfillx {\hspace*{-\textwidth} \hfill} 
\usepackage{natbib}
\usepackage{booktabs}
\ifpdf%
\usepackage{epstopdf}%
\else%
\fi
\usepackage{comment}
\begin{document}
\label{firstpage}

\newcommand{\name}{{\sc ASPER}}

\title{\name: Attention-based Approach to Extract Syntactic Patterns denoting Semantic Relations in Sentential Context}

\author{Md. Ahsanul Kabir\\
	Department of Computer Science\\
	Indiana University Purdue University Indianapolis\\
	\texttt{mdkabir@iupui.edu} \\
	\And
	Tyler Phillips\\
	Department of Computer Science\\
	Indiana University Purdue University Indianapolis\\
	\texttt{phillity@iupui.edu} \\
	\AND
	Xiao Luo\\
	Department of Computer Science\\
	Indiana University Purdue University Indianapolis\\
	\texttt{luo25@iupui.edu} \\
	\And
	\href{https://orcid.org/0000-0002-8279-1023}{\includegraphics[scale=0.06]{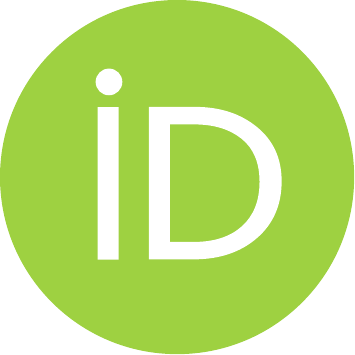}\hspace{1mm}Mohammad Al Hasan}\\
	Department of Computer Science\\
	Indiana University Purdue University Indianapolis\\
	\texttt{alhasan@iupui.edu} \\
}

\maketitle

\begin{abstract}
Semantic relationships, such as hyponym-hypernym, cause-effect, meronym-holonym etc.,\ between a pair of entities in
a sentence are usually reflected through syntactic patterns. Automatic extraction of such patterns
benefits several downstream tasks, including, entity extraction, ontology building, and
question answering. Unfortunately, automatic extraction of such 
patterns has not yet received  much attention from NLP and information retrieval researchers. In this work, we propose 
an attention-based supervised deep learning model, \name, which extracts syntactic patterns between 
entities exhibiting a given semantic relation in the sentential context. We validate the performance of \name\ on three distinct semantic
relations---hyponym-hypernym, cause-effect, and meronym-holonym on six datasets. Experimental results show that for all these 
semantic relations, \name\ can automatically identify a collection of syntactic patterns reflecting the existence of such a relation between a pair of entities in a sentence. In comparison to the existing methodologies of syntactic pattern extraction, \name's performance is
substantially superior.
\end{abstract}

\section{Introduction}

Syntactic patterns within a sentence capture various semantic relationships between the entities in the sentence.
For instance, in a sentence, like, {\tt Sigmoid is a kind of activation function}, the pattern, {\tt $X$ is
a kind of $Y$}, establishes that {\tt sigmoid} and {\tt activation function} share a hyponym-hypernym relationship. Similarly, in a sentence like {\tt COVID-19 causes breathing difficulty in some patients},
the cause-effect relation between {\tt COVID-19} and {\tt breathing difficulty} is reflected by the pattern
{\tt $X$ causes $Y$}. 
Linguists call such patterns {\em syntactic patterns} as they use sentential structures to denote certain relationship between the symbolic feature $X$ and $Y$. 
Extraction of syntactic patterns is an important natural language processing task, as such patterns can
be used to extract entity pairs exhibiting various
semantic relationships~\cite{importance_of_pattern:2018, Volkova:2010:Entity_ext}
and question answering~\cite{jijkoun:2004:information}. Specifically,
 hyponym-hypernym patterns can be used for ontology  building~\cite{klaussner-2011-ontology,Ghadfi-2014-BuildingOF}. Cause-effect patterns
can be used for extracting entities from medical text to discover relations
between disease, symptoms, and medication~\cite{importance_of_pattern:2018,ravikumar2017belminer}.

In existing literature, manual or semi-automatic approaches have been used for extraction of syntactic patterns. Earliest among these works was Hearst's seminal contribution~\cite{Hearst:92, Hearst:1998:WordNetPatterns} on 
finding patterns for hyponym-hypernym relation through manual inspection. Similar manual approaches have also
been used for extraction of patterns denoting cause-effect~\cite{Girju.Moldovan:2002}
and meronym-holonym~\cite{matthew-2002-meronym} relations. But, manual approach for pattern extraction
is laborious and time consuming. Besides, for
every new semantic relationship, an independent inquiry needs to be pursued to obtain  a collection of syntactic 
patterns encoding that relationship.

Snow et al.~\cite{Snow:2005:Automatic} have proposed
one of the earliest semi-automatic syntactic pattern 
extraction method. However, the method is proposed considering
only one kind of semantic relationship, 
hyponym-hypernym. Also, from the methodological 
aspect, the proposed method uses raw frequency threshold of sentential structures over the corpus for selecting a pattern, which generally produces patterns of poor quality.
Subsequent to Snow et al.'s work, another 
semi-automatic work is proposed~\cite{van-2006-part-whole}
for extracting meronym-holonym patterns. This method 
is also based on frequency threshold, and the authors themselves have reported that most of the extracted patterns are false positive. Though the extraction of 
syntactic patterns is not the focus of most of the works, a number of 
works have devoted to utilize syntactic patterns for classifying whether a semantic relationship between a pair of
entities exist or not~\cite{Snow:2005:Automatic, khoo-1998-cause-effect, sorgente2013automatic, sheena-2016-meronym, phi-2016-part-whole}. Note that, extraction of syntactic patterns
is orthogonal to the task of relation classification; 
former extracts syntactic
patterns from the sentences reflecting semantic
relationship, whereas the latter classifies whether a semantic relationship between a pair of
entities exists or not. In this paper our focus is on the former 
task---extraction of syntactic patterns.

Machine learning based methods are also used for predicting 
semantic relation between a pair of entities in a sentence.
Majority of these works~\cite{baroni-lenci-2011-blessed,necsulescu-etal-2015-reading,santus-etal-2015-evalution,santus-etal-2016-nine,shwartz-etal-2016-improving} 
consider the hyponym-hypernym relationship and solve a binary classification problem to identify whether such a
relation holds between a given pair of entities. Such approaches
are often designed to achieve high classification accuracy, but they are not capable of extracting 
syntactic patterns~\cite{Yu:2015:Embedding,sanchez-riedel-2017-well,nguyen-etal-2017-hierarchical}. To summarize,
automatic extraction of syntactic pattern for an arbitrary
semantic relation is yet an unsolved task.

Developing an automated method for extracting syntactic patterns for an arbitrary semantic relation 
is a challenging task. While humans can easily recognize syntactic patterns through a neuro-cognitive process
that enables them to perceive a subject as structured whole consisting of objects arranged in space or sequence,
the same does not hold for a machine learning-based agent, which is better at statistical pattern recognition than 
syntactic pattern recognition. So, it is no wonder that existing computational NLP and AI 
research have not ventured much into the automatic identification of syntactic patterns from natural language text.  Nevertheless, this task is extremely important, because the performance of many NLP applications, such as classification based on pattern
\cite{sorgente2013automatic,Hearst:92,sheena2016pattern,matthew-2002-meronym},
question-answering~\cite{mcnamee-etal-2008-learning, jijkoun:2004:information},
and ontology induction~\cite{poon-domingos-2010-unsupervised} will improve significantly through the automatic 
recognition of such syntactic patterns. 

In this paper, we propose \name \footnote{\name\ is composed of the bold letters in 
{\bf A}ttention-based {\bf S}yntactic {\bf P}attern {\bf E}xtraction for Semantic 
{\bf R}elation}, a generic attention-based deep learning model that can identify 
syntactic patterns for any semantic relationship. 
\name\ follows a supervised learning approach---the 
model is trained through a collection of sentences; for each sentence, an ordered pair of entities 
are identified and a binary label is provided which denotes whether the entities 
are involved into a specific semantic relationship in that sentence. 
The output of the model is a collection of syntactic patterns 
which reflects 
the semantic relationship between entities involving in a chosen semantic relationship.
By changing the training data, in theory \name\ can return syntactic patterns for any semantic
relationship. To obtain the patterns of a given relationship, \name\ uses a bi-directional 
LSTM with an attention layer, which highlights the part-of sentence (pattern) that are 
important to decide whether the identified pair of entities in the sentence are involved 
in  that relationship. Importantly, in the data representation, \name\ does not 
use the embedding vectors of the entities whose relationship is inquired
by the model, which compels \name\ to answer the query by discovering
syntactic patterns capturing that relationship.
Experiments on multiple datasets show \name's effectiveness. 

We claim the following contributions:

\begin{itemize}
    \item We propose \name, a novel deep learning model which can extract syntactic patterns of a chosen semantic relationship between entities in a sentence, effectively and efficiently.
    
    \item Experiments on multiple semantic relationships, such as, hyponym-hypernym, 
    meronym-holonym, and cause-effect show that \name\ can identify most of the previously reported
    syntactic patterns of these relations. It can also identify a few patterns which have not been explicitly noted in  earlier works.
    
\end{itemize}

\begin{figure*}[t]
    \centering
    \includegraphics[scale=0.70]{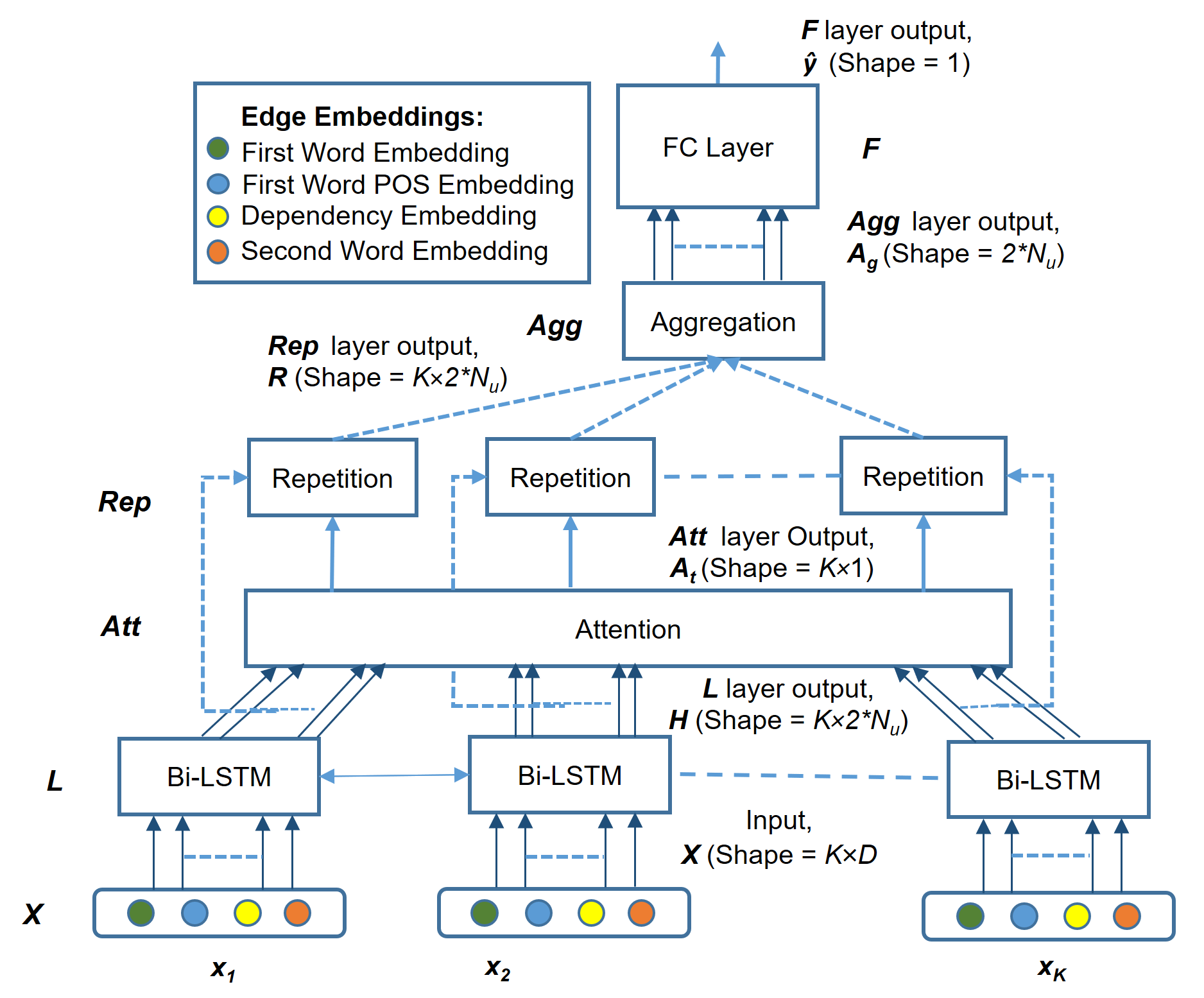}
    \caption{Description of LSTM with attention layer for binary semantic relationship classification.}
    \label{fig:classifier}
\end{figure*}


In this section, we begin by formally defining the relationship-based syntactic pattern extraction task. We then 
describe the LSTM architecture of \name\ along with its input representation and loss function.
Finally, we describe how \name\ extracts syntactic patterns, and provide a pseudo-code of the end-to-end
system.

\subsection{Problem Formulation}
Given a sentence $S$, and a pair of entities (words or phrases) $u, w$ in $S$ exhibiting a specific semantic relationship
$R$ (e.g. hypernymy, co-hypernymy, meronymy, causality, etc.), the task of syntactic pattern extraction is to extract a syntactic pattern, $\mathcal{P}$, which manifests that the entity pairs $(u, w)$ are related through the relation $R$. To extract 
such patterns, in this work, we adopt a supervised learning model. As input, the model takes a set of triplets,
$\mathcal{T} = \{(u_i, w_i), S_i, y_i\}_{i=1}^\Lambda$,  where $(u_i, w_i)$ is a directed pair of entities, $S_i$ is 
a sentence in  which words $u_i$  and $w_i$ co-occur, and  $y_i$ is a binary label indicating if the directed  
entity pair  $(u_i, w_i)$  exhibits the relationship $R$ in the contextual scope of $S_i$; $\Lambda$ is the number
of distinct triples in $\mathcal{T}$. The objective of the 
model is to  extract all syntactic patterns $\mathcal{P}$ such that, $\mathcal{P}$ is associated with one 
or multiple sentences in $\mathcal{T}$ indicating that the entity pairs $(u_i, w_i)$ in those sentences are related through the relation $R$. Note that, a syntactic pattern is a sequence consisting of a subset of linguistic elements 
from the sentences in $\mathcal{T}$, conveying the existence of the corresponding relationship in a 
human-understandable manner. For example, the sentence {\tt LSTM is a type of neural network} exhibits hyponym-hypernym relation between {\tt LSTM}, and { \tt neural network}. The purpose of \name\ is to extract syntactic pattern, {\tt X is a type of Y} from this sentence.

\subsection{Model Architecture}\label{sec:arch}
To successfully extract a syntactic pattern which demonstrates the relationship $R$ 
in entity pair $(u, w)$ in $S$, we must first determine whether $u$ and $w$ exhibit the relationship $R$. To make such distinctions, we train a binary classifier using a supervised
approach through a set of training triples, $\mathcal{T} = \{(u, w), S, y\}$. Since our main objective is to extract syntactic patterns from sentences, a classification model that works with sequential data is needed. In addition, the model should be able to identify the parts of the sentence which contribute the most for making the relationship prediction decision. For these reasons,  we use a bi-directional Long Short-Term Memory (Bi-LSTM)~\cite{Hochreiter:1997:LSTM}  augmented with an attention layer~\cite{Bahdanau:2017:Attention} as our binary classifier. The Bi-LSTM model 
 is able to leverage the sequential nature of our sentence representation. Furthermore, 
as a result of supervised learning, the model's attention layer will be trained to highlight the parts of the sentence that are particularly useful in determining the presence of relationship $R$ 
between the entity pair $(u, w)$. We can, therefore, observe the attention layer to identify the important sentential constructs, which can then be composed to generate the syntactic pattern, $\mathcal{P}$. For a sentence, the Bi-LSTM model takes a vector-sequence 
representation of the sentence and outputs a prediction of the binary label. The complete model is shown in Fig.~\ref{fig:classifier}. 

As shown in the bottom layer of Fig.~\ref{fig:classifier}, the input to the Bi-LSTM is the vector sequence representation of a sentence $S$. This 
representation, denoted as, $\mathbf{X}$, has $K$ edge embeddings in a sequence, each with dimension $D$, 
where the $K$ edges are obtained from the dependency tree of the input sentence. The vector representation of a sentence, composed of a sequence of edge embeddings, is discussed in detail in 
Section~\ref{sec:senrep}.

The Bi-LSTM layer, $\mathcal{L}$, takes $x_{i} \in \mathbf{X}$ as input and outputs two hidden state vectors. The first hidden state vector, $\overrightarrow{h_i}$, is the forward state output, and the second hidden state vector, $\overleftarrow{h_i}$, is the backward state output. Let ${h_i}$ be the concatenated output of  
$\overrightarrow{h_i}$ and $\overleftarrow{h_i}$. Also, we define $\mathbf{H}$, which is the concatenation of 
each $h_i$ output from $\mathcal{L}$ for a single sentence representation $\mathbf{X}$.

\[ \overrightarrow{h}_{i} = \mathcal{L}(\overrightarrow{h}_{i-1},x_{i}),  \overleftarrow{h}_{i} = \mathcal{L}(\overleftarrow{h}_{i+1},x_{i})\]
\[ {h}_{i} = [\overrightarrow{h}_{i}, \overleftarrow{h}_{i}], \mathbf{H} = [{h}_{1}, {h}_{2}....{h}_{K}]\]

Recall that, the shape of a single sentence representation, $\mathbf{X}$, is $K \times D$, where $K$ is the number
of edges in the sentence representation from the dependency tree and $D$ is the dimension of each edge representation.
Therefore, when given a single sentence representation, $\mathbf{X}$, the Bi-LSTM layer, $\mathcal{L}$, produces a concatenated output, $\mathbf{H}$, of shape $K \times 2*N_{u}$, where $N_{u}$ specifies the size of a single hidden vector. For our experiments, we set $N_{u}$  to be equal to 256.

Following the Bi-LSTM layer, $\mathcal{L}$, output $\mathbf{H}$ is used as input to the attention layer, $\bf{Att}$. The attention layer produces, $\mathbf{A}_{t}$, a vector of size $K \times 1$ where each $a_{i} \in \mathbf{A}_{t}$ is a value within a fixed range, $a_{i} \in [0, 1]$. Each such attention value, $a_{i}$, will encode the relative importance of edge embedding $x_{i}$ in making the binary classification decision. $\mathbf{A}_{t}$ is computed as below.

\[\it{Temp} = \bf{Tanh}( \mathbf{H} * \bf{W_{1}} ) * \bf{W_{2}} \]
\[\mathbf{A}_{t} = \bf{Softmax}(\it{Temp})\]

Here $\bf{W_{1}}$ is a trainable matrix of shape $2*N_{u} \times 2*N_{u}$, $\bf{W_{2}}$ is another trainable matrix of shape $2*N_{u} \times 1$. The shape of temporary variable $\it{Temp}$ is $K$, on which we 
apply $\bf{Softmax}$ activation to retrieve $\mathbf{A}_{t}$.

Next, the model uses both $\mathbf{A}_{t}$ and $\mathbf{H}$ as inputs for the repetition layer, $\bf{Rep}$. The repetition layer, $\bf{Rep}$, outputs $\mathbf{R}$ of shape $K \times 2*N_{u}$. $\mathbf{R}$ is simply the scalar multiplication of each hidden input $h_{i} \in \mathbf{H}$ with its corresponding scalar attention value, $a_{i} \in \mathbf{A}_{t}$.

Then, the model uses $\mathbf{R}$ as input for the aggregation layer, $\bf{Agg}$. The aggregation layer simply computes the column-wise sum of $\mathbf{R}$ in order to yield the $2*N_{u}$ shape output, $\mathbf{A}_{g}$. In short, $\mathbf{A}_{g}$ outputs the weighted sum of $\mathbf{H}$ where weights are the attention values. 

\[\mathbf{A}_{g} = \bf{Summation}(\mathbf{R})\]

$\mathbf{A}_{g}$ is then used as input to a fully-connected layer with sigmoid activation function, whose output is a scalar, $\hat{y}$, which denotes the prediction of a binary label, $y$, of a triplet, $t \in \mathcal{T}$.

\[\hat{y} = \bf{Sigmoid}(\bf\mathbf{A}_{g} * {W_{3}})\]

Here $\bf{W_{3}}$ is a randomly initialized weight matrix of shape $2*N_{u} \times 1$.

Using these constructs, we train the binary classifier using the sentence embeddings generated from a collection of triplets, $\mathcal{T}$. We train the model using standard binary cross-entropy loss:
$Loss = -\frac{1}{|\mathcal{T}|} \sum_{t \in \mathcal{T}} y_{t} * log(\hat{y}_{t}) + (1 - y_{t}) * log(1 - \hat{y}_{t})$
Using Early Stopping~\cite{Caruana:2001:EarlyStopping}, we train the model until the validation loss does not decrease at the end of an epoch and then load the model parameters of the previous epoch in which validation has decreased.

\begin{figure*}[t]
    \centering
    \includegraphics[scale=0.4]{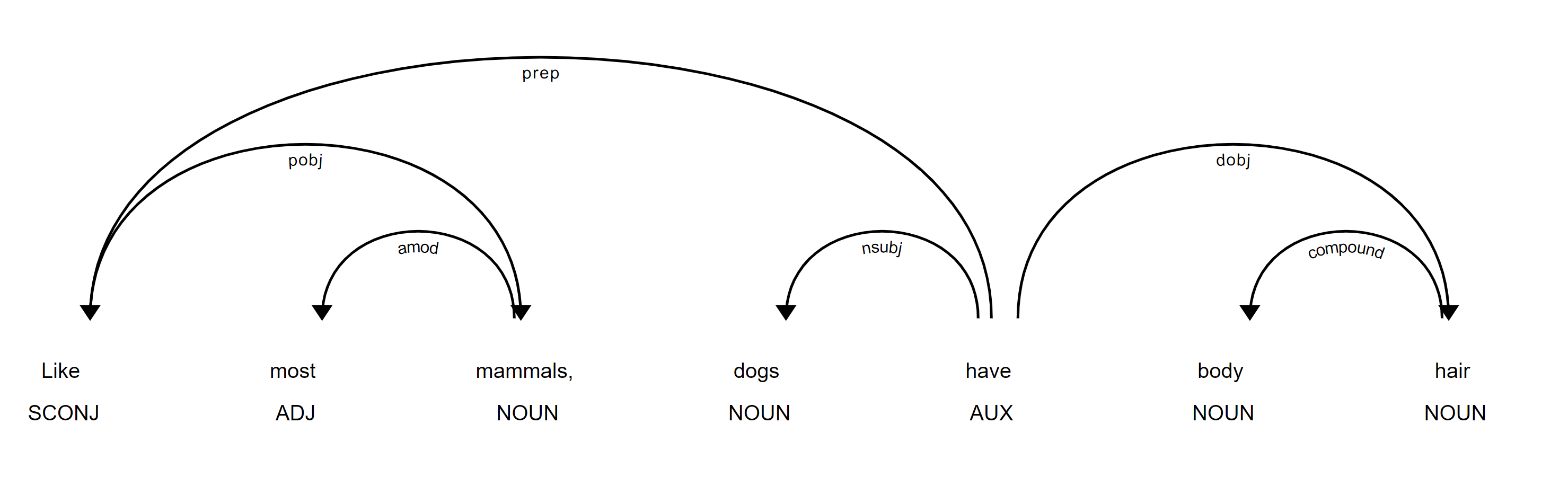}
    \caption{Spacy dependency tree with part-of-speech tags.}
    \label{fig:spacy}
\end{figure*}

\subsection{Sentence Representation}\label{sec:senrep}

To identify syntactic patterns from a sentence using machine learning, the sentence should be
embedded in a form so that its syntactic structure is preserved. For this, we generate a
dependency tree of a sentence~\cite{covington:2001:dep} and use it as input to our 
learning model. The motivation is that the dependency tree of a sentence captures the syntactic 
structure of the sentence through  a parse-tree like structure (see Fig.~\ref{fig:spacy}). However, 
the dependency tree only provides a symbolic representation, so we obtain a vector representation 
of it to be used as input to our model. Given an $N$-word sentence, 
$S$, we obtain a dependency tree of $S$ by using a dependency parser  
(in our implementation we use Spacy 2.2.3~\cite{spacy2} to achieve this). 
Output of this parser is an acyclic directed dependency tree, 
$\mathcal{G} = (\mathcal{V}, \mathcal{E})$. An example of such a dependency tree is given in
Fig.~\ref{fig:spacy}. As shown in this figure, each vertex, $v_{i} \in \mathcal{V}$ is a tuple 
representing a word (or phrase) from $S$ and the part-of-speech tag (e.g. noun, verb, 
adverb, adjective, etc.) of that word ($|\mathcal{V}|=N$).
The edge-set, $\mathcal{E}$, is the set of all directed edges in the dependency tree with cardinality 
$|\mathcal{E}| = M$ $(M < N)$. Each dependency edge $e_{ij}$ links a parent vertex $v_{i}$ to a 
child vertex $v_{j}$, and is labeled by the type of syntactic dependency (attribute, coordinating 
conjunction, compound, etc.) between the words at the two end-vertices of the edge.

In general, the dependency tree of a sentence, $\mathcal{G}$, may contain many vertices and edges which do not 
contribute to conveying if entity pair $(u,w)$ share relationship $R$. For example, consider the sentence: 
{\tt The cat, a type of animal, enjoys laying around and eating}. In this sentence, the first half of the sentence is 
critical in establishing that the {\tt cat} is a type of {\tt animal}. Clearly though, {\tt enjoys laying around 
and eating} plays no role in establishing a semantic relationship between {\tt cat} and {\tt animal}. 
In our sentence representation, we discard such vertices and associated edges.
Specifically, we preserve all vertices and edges which are along the shortest path connecting word pair $u$, and $w$. We also preserve descendants of $u$ and $w$ along with the edges which connect $u$ and $w$ to their descendants.
Next, we organize the edges of the filtered tree into a fixed ordering, in which the edges in the shortest
path between $u$ and $w$ come first, followed by the descendant edges of $u$ and $w$. 

\begin{algorithm}[t]
\caption{Pattern Extraction}
\label{algo:extract_pattern}
\small
\SetAlgoLined\DontPrintSemicolon
    \SetKwFunction{algo}{($\mathcal{T}, Model, att\_thresh, supp, conf$)}
    \SetKwProg{myalg}{Extract\_Pattern}{}{}
    \myalg{\algo}
    {
        $edge\_accum = [ ]$\;
        \For{$t \in \mathcal{T}$}
        {
            $\{(u, w), S, y\} = t$\;
            $\mathbf{X} = SentenceRepresentation(S)$\;
            $\hat{y}, \mathbf{A}_{t} = Model(\mathbf{X})$\;
            \If{$\hat{y} == 1$}{
                $\mathbf{A}_{t} = log(\mathbf{A}_{t})$\;
                $\mathbf{A}_{t} = (\mathbf{A}_{t} - min(\mathbf{A}_{t})) / (max(\mathbf{A}_{t}) -  min(\mathbf{A}_{t}))$\;
                $edge\_index\_mask = \mathbf{A}_{t}[\mathbf{A}_{t} > att\_thresh]$\;
                $edge\_accum.append(S[edge\_index\_mask])$\;
            }
        }
        $\mathcal{P} = ECLAT(edge\_accum, supp, conf)$\;
        $\Return \ \mathcal{P}$\;
    }
\hspace{-0.6in}
\end{algorithm}

In Fig.~\ref{fig:spacy}, the dependence tree of sentence, {\tt Like most mammals, dogs have body hair} is shown. All the edges along the the shortest path from {\tt mammal} to {\tt dog} is part of \name's sentence representation. However, {\tt most}, which is the descendant of {\tt mammals}, is part of the desired syntactic pattern, {\tt Like most 
Y, X}. That is why the descendants of $X$ and $Y$ are also important. But, neither all words on the shortest path, nor the descendants are part of the pattern. That's why \name\ is an attention based approach, so that the important words and edges for patterns can be extracted.

To generate representation of a sentence, we embed each of the selected edges in sorted order and compose 
the resulting ordered  edge representations, $x_{k}$, into a single vector sequence representation of $\mathcal{S}$, $\mathbf{X}$. The
embedding of an edge $e_{k}=(v_i,v_j)$, $x_{k}$, is composed of the following: 
(1) semantic embedding of word (or phrase) corresponding to parent vertex $v_{i}$,
(2) one-hot encoding of part-of-speech tag corresponding to word (or phrase) corresponding to parent vertex $v_{i}$, 
(3) one-hot encoding of syntactic dependency between $v_{i}$ and $v_{j}$, and
(4) semantic embedding of word (or phrase) corresponding to child vertex $v_{j}$.
Zero vectors of appropriate dimension are used for the semantic 
embedding of both the entities $u$ and $w$. This forces \name\ to use only syntactical structural information entailing from sentence 
structure for predicting the relation between
$u$ and $w$, ignoring semantic information from these entity pairs.
For the words except $u$ and $w$, we use 512-dimensional universal
sentence encoder (USE) vectors~\cite{cer-etal-2018-universal}. Note that, one may use other choices,
such as word2vec, or Glove, instead of USE. For the
part-of-speech tags and syntactic dependencies types,  we use one-hot encoding, using 18-dimensional 
and 58-dimensional vectors, respectively. Therefore, any edge embedding has a fixed dimension $D$, 
which is equal to $512+512+18+58 = 1100$. Finally, we fix vector sequence $\mathbf{X}$ to a fixed-length $K$ (the number of edges) by either removing edge embeddings from the end of the sequence or adding 
zero-padding vectors of size $D$. This ensures that any sentence representation, $\mathbf{X}$, is of a 
fixed size, $K \times D$. Clearly, our sentence representation, $\mathbf{X}$, is agnostic to the relationship 
$R$, so it is capable of encoding an arbitrary semantic relationship between a given entity pair $u$ and $w$.

\subsection{Pattern Extraction Pipeline and Pseudo-code}

After training the supervised learning model as discussed in Section~\ref{sec:arch}, the model can be
used for classifying whether an unseen pair of entities (within the context of a sentence) shares a
relationship or not. This works for an arbitrary semantic relationship as long as we can gather
training data for that relationship. However, our main aim is automatic extraction of syntactic 
patterns, so we consider each edge in an edge-set as an item and apply frequent itemset mining algorithm
ECLAT~\cite{Zaki:2000:Eclat} to obtain frequent edge-sets over the sentences of $\mathcal{T}$, which
constitutes the desired syntactic patterns.

The pseudo-code of \name\ is given in Algorithm~\ref{algo:extract_pattern}.
For a triplet $t (u, w, S)$ in a given collection $\mathcal{T}$, we first train the model and predict the label
$\hat{y}$ and the attention values, $\mathbf{A}_{t}$, associated to the edges of $t$ (Line 3-6). Next, 
we consider each $t \in \mathcal{T}$, for  which the model predicts positively i.e., confirms the 
existence of  relationship $R$ in entity pair $(u, w)$ in the sentence $S$ (Line 7) . For
the qualified triples, we normalized the attention values of their edges, and by using an importance
threshold, $att$ (a value between 0 and 1), filter out the edges of lesser relative importance (Line 10). As the attention values are 
on an exponential scale (output of a Softmax function), before applying the threshold, we take the logarithm 
of the attention values and then use min-max normalization to scale the attention values between 0 
and 1 (Line 8-9). Corresponding to each triple, we accumulate an edge-set considering only the important
edges (Line 11). Then frequent pattern mining algorithm is used to obtain syntactic pattern-set (Line 14). 

\begin{table*}[t]
\small 
\centering
\caption{Dataset Statistics}
\begin{tabular}{c c c c c c c} \toprule
    \bf Dataset & \bf Train Pairs & \bf Val Pairs & \bf Test Pairs & \bf Train Sentences & \bf Val Sentences & \bf Test Sentences \\
    \midrule
    LEX  & 20,335 & 1,350 & 6,610 & 104,117 & 1,305 & 33,701 \\
    RND & 49,475 & 3,534 & 17,670 & 236,859 & 3,435 & 89,883 \\
    \midrule
    SemEval & 6,914 & 1,053 & 2,838 & 7,157 & 1,166 & 2,861 \\
    ADE & 7,917 & 1,625 & 6,340 & 8,162 & 1,636 & 6,503 \\
    \midrule
    BLESS & 11,151 & 3,225 & 10,163  & 23,412  & 3,889 & 19,512 \\
    Phi & 4,638 & 812 & 2,853 & 7,938 & 1,587 & 6,352  \\
    \hline
\end{tabular}
\hspace{-0.1in}
\label{tab:dataset-statistics}
\end{table*}
\section{Experiments and Results}
As \name\ is relation-agnostic, we validate its performance in extracting 
syntactic patterns for multiple relations; specifically, 
we choose hyponym-hypernym, cause-effect and meronym-holonym relationships, as these three are well-studied semantic
relations in the literature. We also compare the performance of \name\ with
Snow's method~\cite{Snow:2005:Automatic}, the only semi-automatic method (to be best of our knowledge) that
extracts syntactic patterns. However, Snow's method works only for the hyponym-hypernym
relation, so we compare with this method for results on this relation.
For the other relations that we experimented with, we are not aware of a method, barring from manual methods~\cite{Girju.Moldovan:2002,matthew-2002-meronym}, so for these
relations, we show results on \name\ only.

\subsection{Datasets}\label{sec:dataset}

We use six datasets for showing the performance of \name. The statistics of the datasets
are shown in Table~\ref{tab:dataset-statistics}. Among these, LEX, RND are used for 
hyponym-hypernym pattern extraction; SemEval, ADE datasets are used to perform cause-effect 
pattern extraction, and the remaining, Bless and Phi's datasets are used for meronym-holonym 
pattern extraction tasks. 

Our problem formulation requires context sentences for the entity pairs, but four of the six
datasets do not have any context sentence associated to the entity-pair. We obtain context
sentences from Wikipedia. For this, we download the latest wikipedia dump and  
extract all the sentences. Then, if a pair of entities co-occur in a sentence, we extract and
associate that sentence with the entity pair. Note that, in this way, a given pair can be associated
to multiple sentences. Also, important to understand that not every sentence has a pattern even
if the sentence contains an entity-pair. On some occasions, sentences merely list a pair of entities,
but do not imply a relationship between them in the sentential context. The pattern extraction methods
(\name\ and Snow's method) only extract patterns if the model predicts that a relation between a pair 
of entities exists in the  sentential context. More details of these datasets are provided below.

\noindent {\bf LEX \& RND:} These datasets are obtained from ~\cite{shwartz-etal-2016-improving}.
They list a set of entity pairs with a label denoting whether the entity pair have a hyponym-hypernym
relation (positive) or not (negative) without context sentences for an entity-pair. As discussed above,
we use Wikipedia for obtaining context sentence for an entity-pair. Since multiple Wikipedia sentences 
can be associated to a given entity pair, for both the datasets, we allow at most five sentences to be
associated to an entity pair. Both LEX and RND datasets are balanced having the same number of positive 
and negative sentences. Also, these datasets are already split into  train, test, and validation partition
which we respected. In LEX dataset, disjoint entity pairs are used in train and test partition; while RND is 
split randomly, so same entity pair may appear in train, validation, and test partitions, but with distinct sentences.

\noindent {\bf Bless:} We use this dataset for evaluating meronym-holonym pattern extraction. It was used in
\cite{baroni-2011-bless} for classifying different semantic relationships. It does not
have any context sentence, so we extract sentences from Wikipedia
for these pairs. Since this dataset has entity pair for many relations, we consider
meronym-holonym entity pairs as positive class and others as negative class.
For both positive and
negative entity pairs, we allow at most 3 sentences for each pair. Finally, we maintain positive and negative sentence ratio as 1:1; split the dataset into train, test and validation maintaining 50\%, 40\%, 10\%, 
respectively.

\noindent {\bf Phi:} This dataset is used in this paper \cite{phi-2016-part-whole}, in which
authors (Phi et al., whose name is used for naming this dataset) used word embedding
for extracting different kinds of meronym-holonym relationship between entities. We 
use this dataset for evaluating meronym-holonym pattern extraction. This dataset contains only positive pairs 
with different kinds of part-whole relationships, such as, component-of (11.2\%), member-of (22.21\%), stuff-of (18.89\%), participates-in (15.23\%), etc. as labels. For the negative pair sentence instances, we borrow from Bless dataset. We maintain positive negative sentence ratio as 1:1 so that the dataset is balanced. Finally, We split the dataset randomly for train, test and validation partitions maintaining 50\%, 40\%, and 10\% instances respectively.

\noindent {\bf SemEval:} This is a well-used dataset, built by combining  the SemEval 2007 Task 4 dataset~\cite{girju-etal-2007-semeval} and the SemEval 2010 Task 8 dataset~\cite{hendrickx-etal-2010-semeval}. The SemEval datasets provide predefined positive and negative sentences with corresponding entity pairs. The datasets also include predefined train and test partitions. For building validation partition, we borrow from the train
partition. Unlike the previous datasets, this dataset is imbalanced. The ratio of positive and negative sentences is 1:5. And the ratio of train, test, and validation sentences is 7:3:1.

\noindent {\bf ADE:} The adverse drug effect (ADE) dataset~\cite{Gurulingappa-2012-ADE} includes a set of predefined positive and negative examples. In the case of the positive examples, the cause-effect entity pair is given along with a corresponding sentence. On the other hand, the negative examples are only a collection of sentences that do not exhibit the cause-effect semantic relationship. In order to have entity pairs for each negative sentence, we randomly obtain two noun phrases from each negative sentence. This dataset is balanced in terms of number of sentences and train, test, and validation partitions contain 50\%, 40\% and 10\% data respectively.


%


\begin{table*}[t]
\caption{Pattern Extraction Results of ASPER: Evaluated on Sentence(Left), Evaluated on Pattern(Right)}
        \centering
        \begin{minipage}{0.3\textwidth}
            \scalebox{1}{
                \begin{tabular}{ l  l  c  c  c }
                \bf Relation &  \bf Dataset & \bf Prec & \bf Rec & \bf F$_1$ \\
                \toprule
                  \bf Hyponym- & \bf Lex & 0.78  & 0.7 & 0.74 \\
                  \bf Hypernym & \bf RND & 0.88  & 0.72 & 0.80 \\
                  \midrule
                  \bf Meronym-  & \bf Bless & 0.52  & 0.58 & 0.54 \\
                  \bf Holonym & \bf Phi & 0.62  & 0.67 & 0.64 \\
                  \midrule
                  \bf Cause- & \bf ADE &  0.69  & 0.61 & 0.65 \\
                   \bf Effect & \bf Semeval &  0.71  & 0.71 & 0.71 \\
                \hline
                \end{tabular}
            }
        \end{minipage}
        \hfillx
        \centering
        \begin{minipage}{0.46\textwidth}
            \scalebox{1}{
                \begin{tabular}{ l  l  c  c  c }
                \bf Relation &  \bf Dataset & \bf Prec \\
                \toprule
                \multirow{2}{*}{\bf Hyponym Hypernym} & \bf Lex & 0.82 \\
                 & \bf RND & 0.87 \\
                  \midrule
                 \multirow{2}{*}{\bf Meronym Holonym}  & \bf Bless & 0.55 \\
                 & \bf Phi & 0.64 \\
                  \midrule
                  \multirow{2}{*}{\bf Cause Effect}  & \bf ADE &  0.68  \\
                & \bf Semeval &  0.73  \\
                \hline
                \end{tabular}
            }
        \end{minipage}
        \label{tab:pattern-extraction-method}
\end{table*}

\subsection{Hyper-Parameters Discussion}

For training  \name,  we first utilize LSTM model for identifying important dependency tree edges which
later constitute the syntactic patterns. This step involves a few user-defined parameters: $K$ (the maximum 
number of dependency tree edges in the  sentence representation), $batchSize$ (total number of train instances 
in a batch), the size of hidden layer ($N_u$) in a Bi-LSTM unit, and the learning rate. We fix the hidden layer 
size at 256, without tuning. We use Adam optimizer with its default learning rate (0.01), and early stopping.
We tune $K$ from the value between 30 and 40, and tune $batchsize$ from the values $\{128, 256, 512\}$.
For LEX, RND, SemEval, ADE, BLESS and Phi datasets, we find the best results for 
$K$ = 30, but for the ADE, $K$ = 35. For all datasets, $batchSize$ equal to 128 produces the best result.

For extracting patterns from the important dependency edges by using frequent
itemset mining, we have two hyper-parameters, $supp$ (minimum support threshold 
in percentage), $conf$ (minimum confidence threshold in percentage). Another
hyper-parameter is $att$ (Attention threshold) which is used to filter the important edges. $att$ is tuned for the values between 
0.1 to 0.9 at 0.1 interval. We get good patterns for $att = 0.6$ for all the datasets except Phi where $att = 0.1$ works well. Both $supp$ and $conf$ are tuned using a validation set from values 
between  0.1\% to 3.0\% at 0.1 interval; the patterns that we obtain from validation set are 
manually scanned to choose the optimal values of $supp$ and $conf$. For small value of these 
parameters, we find noisy and incomplete patterns, which do not qualify as syntactic patterns of a 
relation. Alternatively, if those values are too large, we find too few patterns. We find 
that small support and confidence threshold work the best as they obtained larger patterns, denoting 
a full syntactic pattern, conveying a semantic relationship. For LEX and RND optimum $supp$ values 
are 0.28\% and 1.3\%; optimum $conf$ values are 0.7\% and 0.7\%. For the SemEval Combined, ADE, 
optimum $supp$ values are 0.3\%, 0.5\% and the optimum $conf$ values are 0.5\%, 0.4\%. Finally, 
for Bless and Phi datasets $supp$ values are 0.3\%, 1.0\% and the optimum $conf$ values are 
0.4\%, 0.5\% respectively. We perform ablation study over $supp$ (results shown in Section~\ref{sec:ablation}).

\subsection{Pattern Evaluation}
Evaluating a pattern extraction is a difficult task as the ground truth for a pattern extraction method
is not available. Existing works, manual or semi-automated only perform a qualitative evaluation. In this
work we have proposed two quantitative metrics for evaluating the performance of pattern extraction. We
discuss them below.\\

\noindent{}{\bf Evaluation on Sentence}\\
Our first evaluation method builds ground truth by manually extracting patterns directly from 
the sentences in a dataset. Unfortunately, such an effort is time consuming and difficult for large datasets. 
So such an evaluation is only possible by sampling a subset of sentences
in a dataset. So, given a potentially large test dataset, we first choose a
random subset of sentences (around 1000) from the positive class (where the entity pair in the sentence 
exhibit the  relation). For each of these sentences, we manually extract the pattern and make 
a ground truth pattern set over a sample of the dataset. If $\mathcal{P}_t$ is the total 
pattern set and $\mathcal{P}_o$ is the obtained pattern set over the same sentences in the 
sample of the dataset, the following equations define precision and recall of pattern extraction by a method.

    \[prec = \frac{|\mathcal{P}_o\cap\mathcal{P}_t|}{|\mathcal{P}_o|}, rec = \frac{|\mathcal{P}_o\cap\mathcal{P}_t|}{|\mathcal{P}_t|}\]

\begin{figure*}[t!]
    \centering
    \includegraphics[scale=0.55]{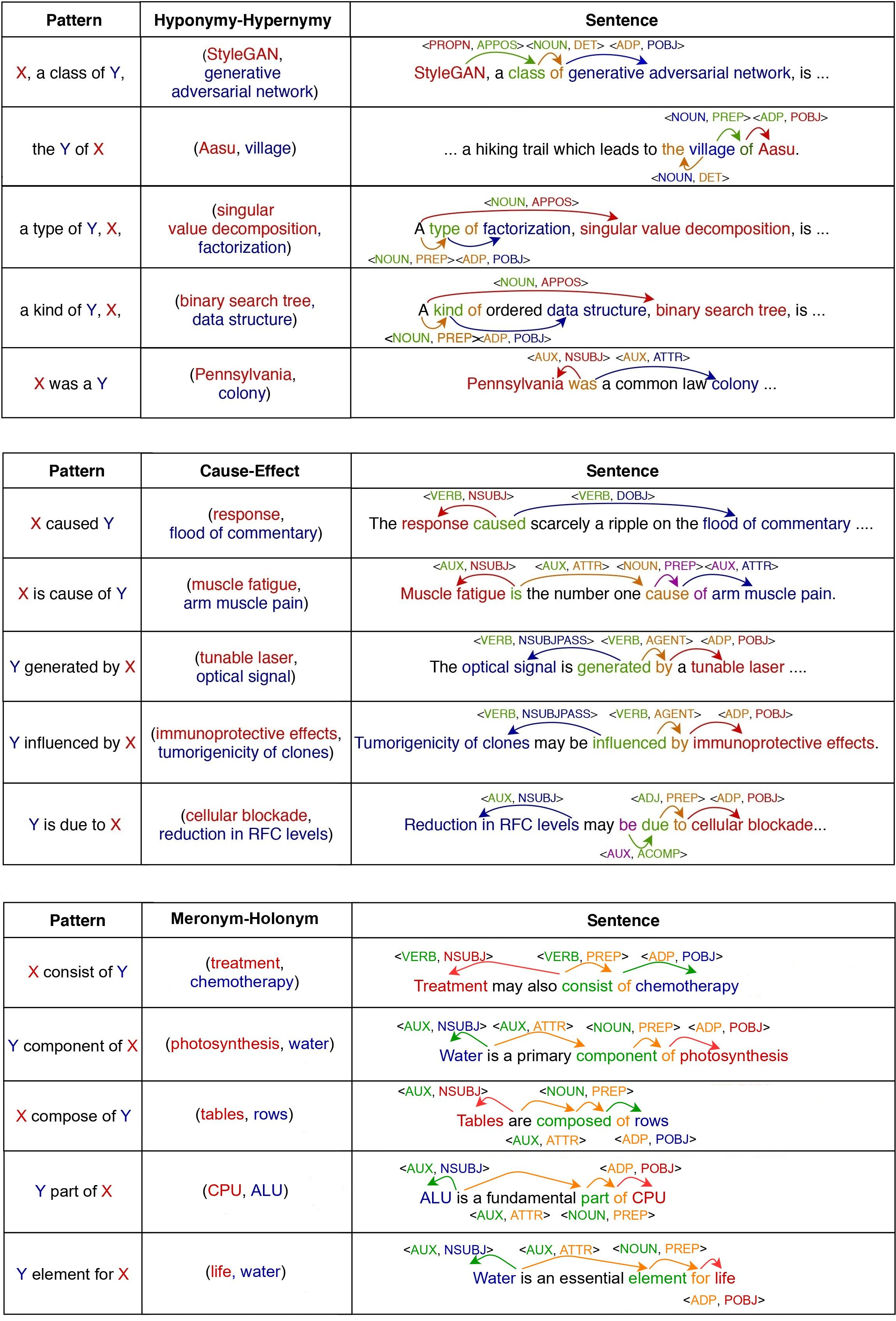}
    \caption{Example extracted syntactic patterns} 
    \label{fig:all-combined-pattern}
\end{figure*}

A problem with the previous evaluation metric is that it is computed over a random
sample of sentences in the dataset, not the entire dataset. In fact, it is impractical to extract pattern 
manually over all the sentences in a dataset. But for any semantic relations, there 
generally exist a finite number of patterns, and it is easier to validate these patterns without
observing them in the sentential context. In this evaluation method we manually evaluate the 
precision of extracted patterns (over the entire dataset) by a method without evaluating them 
in the sentences. In other words, all the correctly predicted patterns in an extracted pattern set 
is considered to be the ground truth, and precision is computed as the ratio of correctly predicted
patterns over all the extracted patterns. If we have more than one pattern extraction methods, we 
collect all the correctly predicted patterns by all of the methods and consider that to be the
ground truth pattern-set and report precision on the basis of this set. Evaluation on pattern is
easier because the number of patterns generally in less than a hundred for a given semantic relation,
and manual evaluation of a pattern is still possible without considering it in the sentential context.
Let $\mathcal{P}_t'$ be set of collected patterns in the ground dataset and $\mathcal{P}_o'$ be the obtained pattern set by a specific method. Then, we define precision 
and recall of the method with similar equations like before.

    \[prec = \frac{|\mathcal{P}_o'\cap\mathcal{P}_t'|}{|\mathcal{P}_o'|}, rec = \frac{|\mathcal{P}_o'\cap\mathcal{P}_t'|}{|\mathcal{P}_t'|}\]
    
However, note that in this kind of evaluation, a method is not penalized for not discovering
a pattern as long as no other competing methods is able to discover that pattern.

\subsection{Quantitative Pattern Extraction Results}

In this section we first discuss the performance of \name\ for its ability to extract
patterns for three distinct relationships: Hypernym-Hyponym, Cause-Effect, and Meronym-Holonym, from six datasets, two for each relation. In 
Table \ref{tab:pattern-extraction-method}, we show the results 
for all the datasets for the precision, recall, and $F_1$ metrics using
sentence based evaluation. Over all the datasets and various relations, \name's
performance is the best for detecting patterns for Hyponym-Hypernym relation with $F_1$
score of 0.74 and 0.88 on Lex and RND datasets, respectively. The poorest performance
was for the Meronym-Holonym pattern with 0.54 and 0.64 $F_1$ score on two of its datasets.
The reason for the best performance for Hyponym-Hypernym relation is possibly due to
well-established patterns for expressing this relation in a sentence. For the other
two relations, the syntactic patterns are more fluid and hence, hard to recognize
by an automated method. 
That means, even if a pair holds a semantic relation, only a few sentences have a syntactic pattern. We observe this from sampled test dataset which is labelled manually. Similar argument holds for ADE cause-effect dataset.  The performance on Semeval dataset is comparatively better. This dataset was created for competition and many of the sentences
in this dataset are constructed with true cause-effect patterns. 
Finally, although there are already sentences for Phi dataset, the sentences do not 
always have consistent syntactic patterns.

\begin{figure*}[t!]%
    \centering
    \subfloat[][\centering Pattern Extraction Results in Lex Dataset]{\includegraphics[width=6.2cm, height=4.8cm]{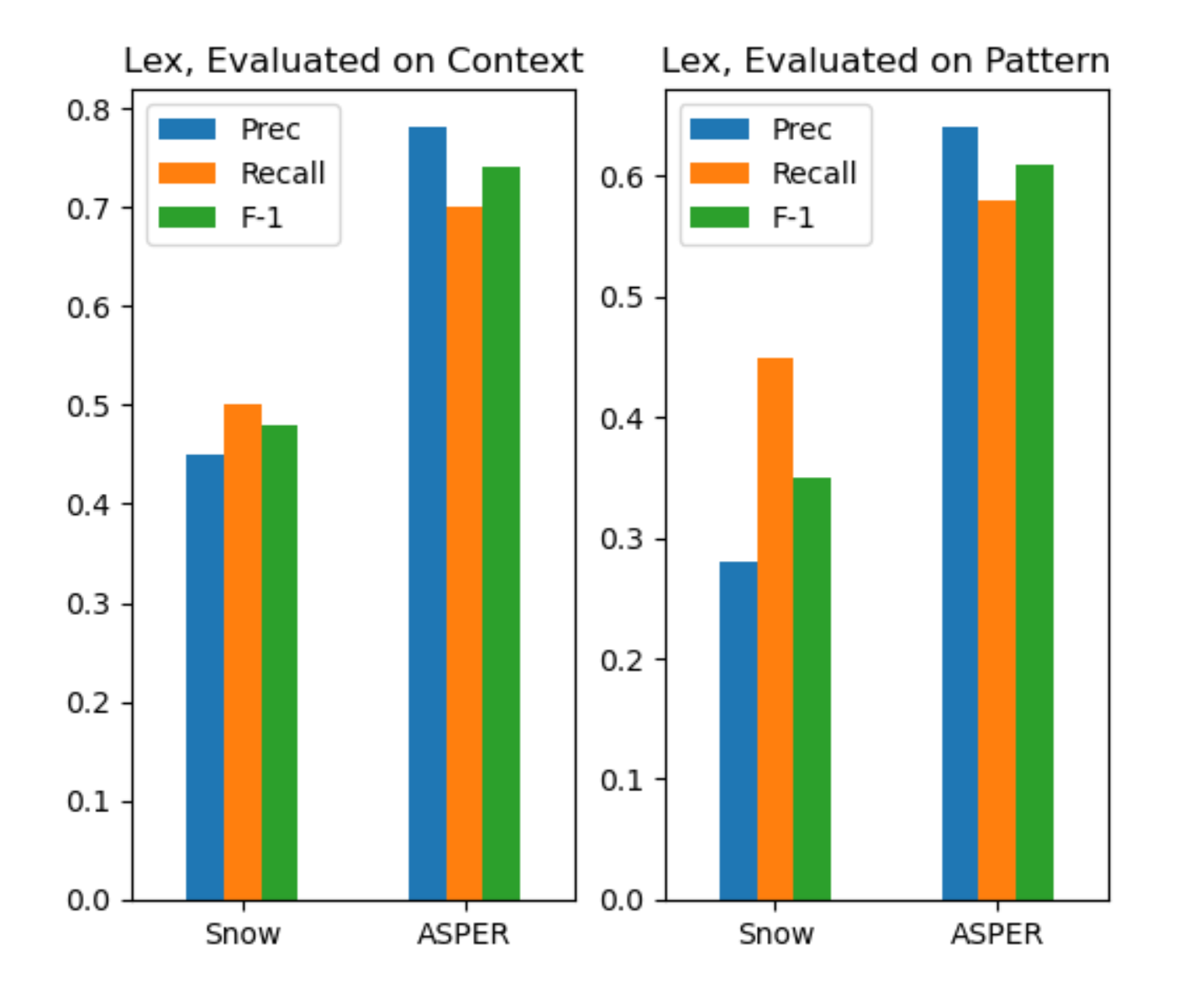}}
    \subfloat[][\centering Pattern Extraction Results in RND Dataset]{\includegraphics[width=6.2cm, height=4.8cm]{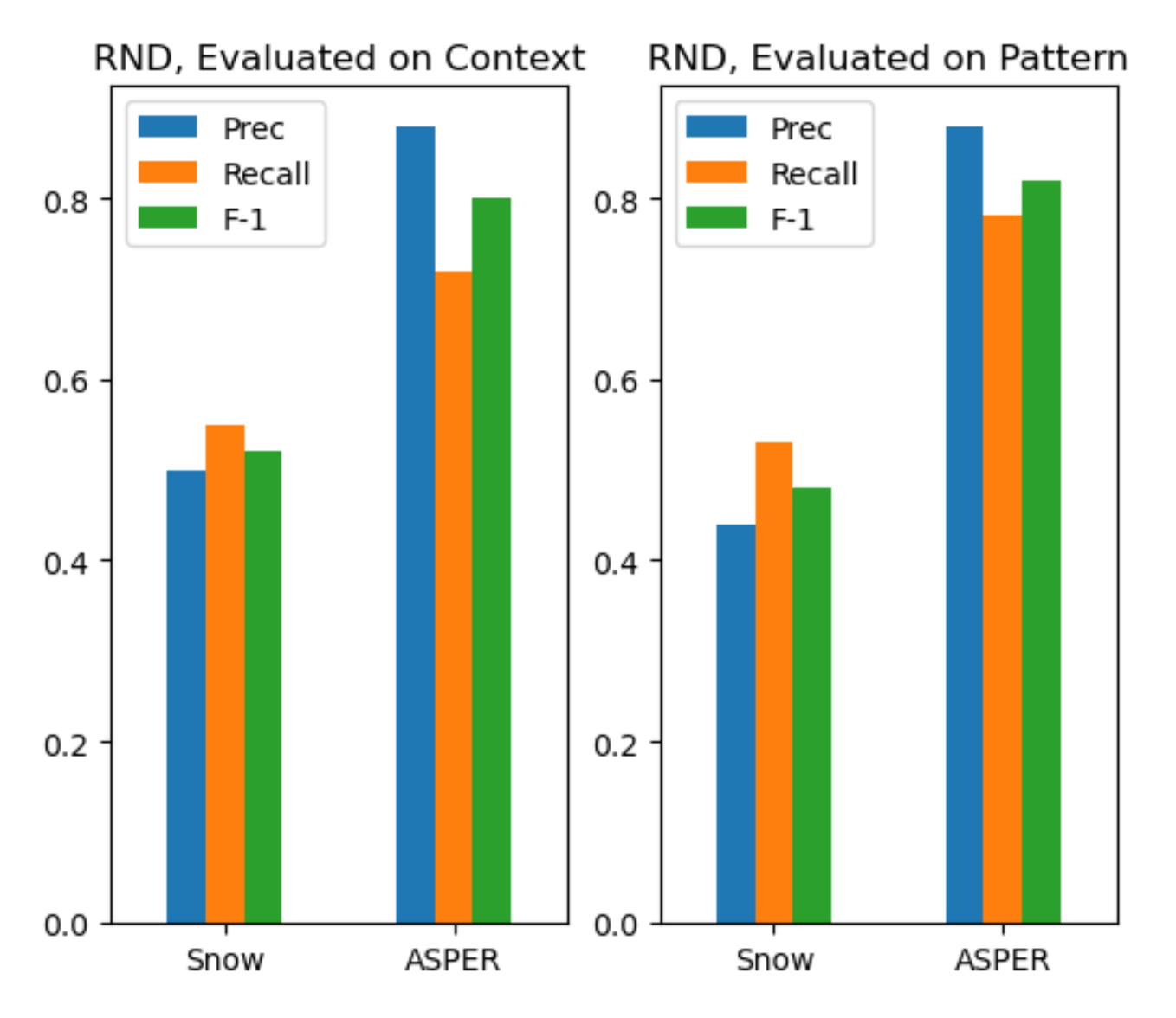}}
    \caption{Hyponym-Hypernym Pattern Extraction Results}%
    \label{fig:hypo-hypr-result}%
\end{figure*}

In Table~\ref{tab:pattern-extraction-method} we show the results by using the 
evaluated on pattern approach. The finding is very similar to the results in 
Table~\ref{tab:pattern-extraction-method}. Note that for the evaluation metric 
based on pattern, only precision is shown. This is due to the fact that for
pattern based evaluation when only one extraction method is used, we have no knowledge
about false-negative, so recall cannot be computed.

We compare the performance of \name\ with that of Snow et. al~\cite{Snow:2005:Automatic} work, 
the latter works only for Hyponym-Hypernym pattern extraction. So we show
comparison results on Lex and RND datasts for the Hyponym-Hypernym pattern
extraction task. This comparison
result is shown in the bar chars of Figure~\ref{fig:hypo-hypr-result} 
using precision, recall and $F_1$ values of both the pattern evaluation metrics. 
Both the methods are tuned for the highest $F_1$ score.
As we can see from the bar chart,
for both the datasets (Lex on the Left, RND on the right), with respect to both evaluation metrics, 
\name\ beats Snow's method significantly.
In fact, precision, recall, and $F_1$ of Snow's  method are substantially lower (50\% lower) than 
\name\ for both the evaluation metrics in both the datasets.

Although we could not compare \name\ with other methods for meronym-holonym and cause effect patterns extraction, the results in Table \ref{tab:pattern-extraction-method} and \ref{tab:pattern-extraction-method} clearly
indicate that, \name\ performs well for extracting patterns for other relations.

\begin{figure*}[]%
    \centering
    \subfloat[][\centering RND Ablation]{\includegraphics[width=4.5cm, height=4.5cm]{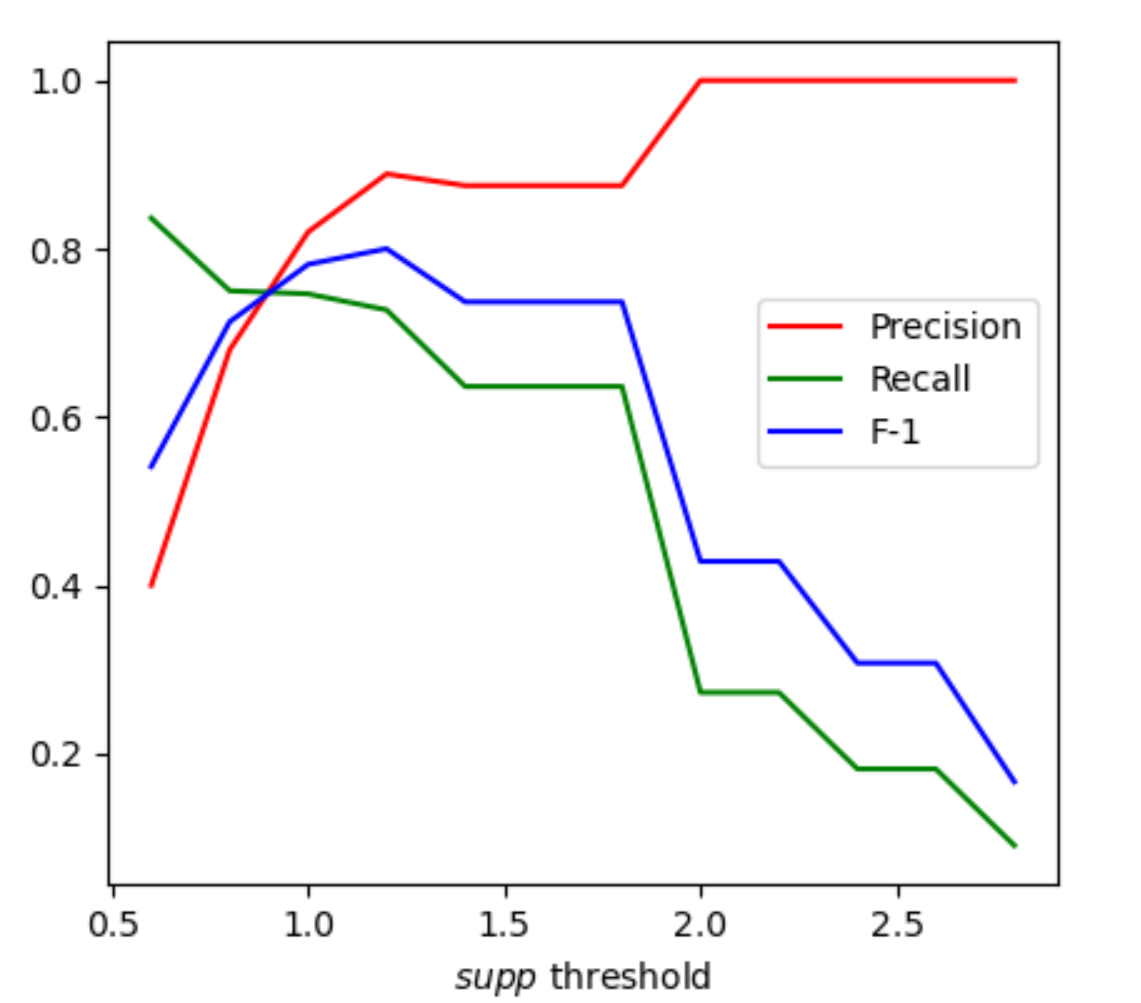}}%
    \subfloat[][\centering ADE ablation]{\includegraphics[width=4.5cm, height=4.5cm]{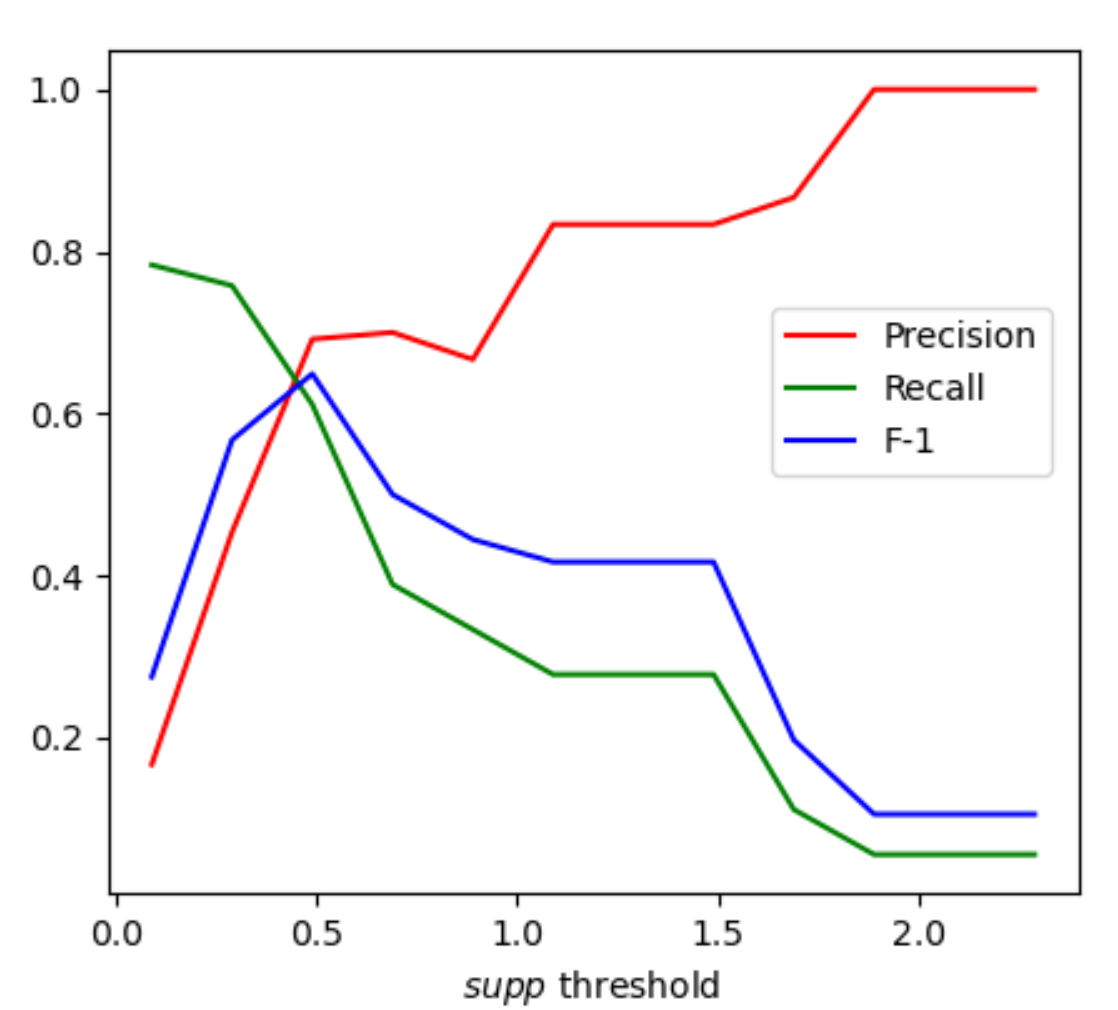}}%
    \subfloat[][\centering Phi ablation]{\includegraphics[width=4.5cm, height=4.5cm]{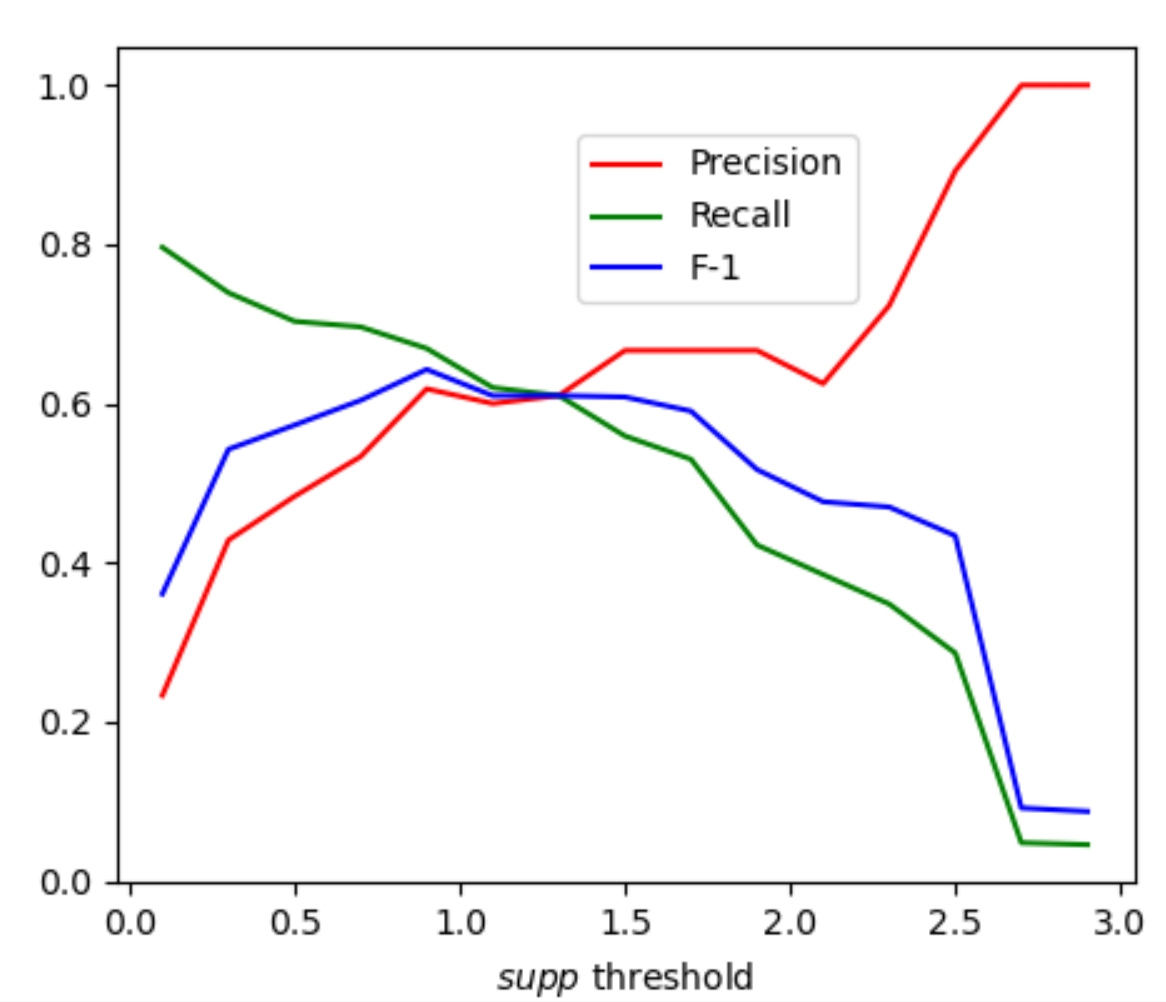}}%
    \caption{Ablation Study with ASPER}%
    \label{fig:ablation-study}%
    \hspace{-1.0in}
\end{figure*}
\subsection{Qualitative Pattern Extraction Results}
In Figure \ref{fig:all-combined-pattern}, we show some of the patterns extracted by \name. The
first column shows the human readable patterns; the second column shows the entity pairs which exhibit the semantic relationship, and finally the third column shows the sentences with the corresponding semantic pairs and the syntactic dependency patterns extracted by \name. 
For example, the second row of Hyponym-Hypernym patterns in Figure~\ref{fig:all-combined-pattern}, contains the pattern {\tt the Y of X}.
The word {\tt Aasu} is a hyponym of the hypernym, {\tt village}. For this hyponym-hypernym pair, the sentence {\tt A hiking trail leads to the village of Aasu} is extracted from Wikipedia from which \name\
identifies the dependency edges {\tt village} $\rightarrow$ {\tt the}, {\tt village} $\rightarrow$ {\tt of} and {\tt of} $\rightarrow$ {\tt Aasu}; from the attention values and itemset mining. If we replace {\tt Aasu} with $X$, and {\tt village} with $Y$
we get the hyponym-hypernym pattern, {\tt the Y of X} in the first column, which is not reported by Hearst and Snow \cite{Hearst:92,Snow:2005:Automatic}. Along with finding new hyponym-hypernym patterns, \name\ re-discovers most of the Hearst patterns. Similarly, the third row of cause-effect patterns in Figure~\ref{fig:all-combined-pattern}; {\tt Y generated by X} is a pattern which is not reported by \cite{khoo-1998-cause-effect}, and the fourth row, {\tt Y influenced by X} is not used by \cite{sorgente2013automatic} for classification. For meronym-homonym, {\tt Y element 
for X} is not used by \cite{sheena-2016-meronym}.


\subsection{Ablation Study}
\label{sec:ablation}

The main hyper-parameter of \name\ which affect its pattern extraction performance are $supp$, and $conf$.
We show \name's performance over varying $supp$ values by keeping $conf$ value fixed for one dataset for
each of the relations. The trend with varying $conf$ is similar to $supp$'s variation, hence not shown.
The findings are shown in Figure~\ref{fig:ablation-study}. In this Figure, for each plot, support values
are shown along the $x$-axis and performance values (precision, recall, $F_1$) are shown along the $y$-axis.
From all the three plots, the $F_1$-score values increase as $supp$ increases reaching at the peak, then gradually decreases. With larger $supp$ precision always increases, as with higher support more stringent requirements is imposed for the selection of a pattern. On the other hand, the recall curves always go downward direction since the number of predicted patterns decreases as $supp$ increases.

\section{Related Works}\label{sec:related}

Related works are discussed in two groups. The first group comprises the works which do not 
extract syntactic patterns, rather perform semantic relationship classification.  A subset of 
these works first manually collect patterns and then use them for pattern-based 
semantic relationship classification
\cite{sorgente2013automatic,Hearst:92,sheena2016pattern,matthew-2002-meronym}; these
methods can be benefited by the availability of an automated pattern extraction tool, like \name.
Other works in this group use embedding vectors, entities, dependency edges etc.\ followed by 
supervised classification of relationship~\cite{li2019knowledge, wu-2019-bert-classification, lee-semantic-relation-classification-2019, Yu:2015:Embedding,nguyen-etal-2017-hierarchical,sanchez-riedel-2017-well,shi-2019-bert-classification, xu-2019-neural-relation, shwartz-etal-2016-improving,shwartz-etal-2017-hypernyms, zhang-2018-graph-convolution, shwartz-etal-2017-hypernyms}. Some of the recent works among these 
use deep learning with attention for classification~\cite{Wang:2016:Classification, shen-huang-2016-attention,Ji2017DistantSF}. However, the attentions are not used to mine syntactic pattern,
rather to validate whether the model is concentrating over the important segment of the sentence.

Works in the second group explicitly focus on automatic pattern extraction, but each of the work
is limited to one specific semantic relationship. For instance, \cite{Snow:2005:Automatic} method
(which we compare against) first obtain dependency trees, and then apply a frequency threshold over
the edges to obtain patterns to obtain hyponym-hypernym patterns. Generally, frequency-based methods 
generate a large number of noisy and false positive patterns. For meronym-holonym relation a google 
search based semi-automatic, frequency based approach exists in the literature \cite{van-2006-part-whole}. 
The authors reported to find 1000 snippets, and 4503 unique patterns for 503 part-whole pairs. Top 300 
frequent patterns out of 4503 patterns are manually validated and they claim to get only 12 correct 
patterns.



\section{Conclusion}
We present \name, a novel deep learning model which can extract syntactic patterns shared between
entity pairs within a sentential context to convey a semantic relation. It works for any relation, it can predict 
the existence of a relation, and it  can also extract syntactic patterns of that relation---a unique feature that no 
existing method can offer. 
The experimental results show that \name\ can extract all known syntactic patterns of a relation, 
including a few new patterns which are not explicitly stated in the previous works. 
The future work of this research includes applying \name\ to extract syntactic patterns of other semantic relationships. Another research goal is to use the patterns,
specifically cause-effect patterns to extract entity-pairs showing relationships between
disease, symptoms, and medication.
Authors are committed to reproducible research, and will
release code, and ground truth datasets once the paper is accepted.

\bibliography{cam2021,anthology.bib}

\onecolumn
\appendix

\section{Appendices}
Table ~\ref{hypo-patterns-appendix} shows all the hypoonym-hypernym patterns extracted by \name. However, the first three patterns are same in terms of formation, only the semantic is different. The same sentence is paraphrased to show the similarity of these three patterns. However, the other patterns are kept as it is. Table ~\ref{cause-effect-appendix} and ~\ref{mero} show the cause-effect and meronym-holonym patterns respectively.
\label{sec:appendix}
\setcounter{table}{0} \renewcommand{\thetable}{A.\arabic{table}}

\begin{table}[h!]
\footnotesize
\caption{All Meronym-Holonym patterns extracted by \name}\label{mero}
\setlength{\tabcolsep}{1.5mm}
{\begin{tabular}{p{0.18\linewidth}|p{0.15\linewidth}|p{0.15\linewidth}|p{0.4\linewidth}}\toprule
\bf Pattern & \bf X & \bf Y & \bf Sentence \\
\midrule
 X consist of Y \newline
 X comprise of Y
 &  treatment
 & chemotherapy
 & Treatment may also consist of chemotherapy

 \\\hline
 
 Y part of X \newline
 Y element for X\newline
 Y source of X \newline
 Y component of X \newline
 Y constituent of X 
  &   Lowe Group
 & Lowe
 & Lowe is part of the Lowe Group, one of the three large subsidiaries of Interpublic.
 \\\hline
  
  X made of Y 
  &   Gene
 & DNA
 & Genes are made of DNA

\\\hline

Y block of X \newline
  &   Muscle
 & Protein
 & Protein is the building block of muscle
 
 \\\hline
 X have Y \newline
  &   The commission
 & seven members
 & The commission shall have seven members.

 \\\hline
 X group of Y \newline
  &   arthropods
 & invertebrates hypoxaemiant pneumonitis
 & Arthropods are a group of invertebrates.
 
 \\\hline
 
X mixture of Y 
  &   Concrete
 & cement
 & Concrete is a mixture of cement.

 \\\hline
 X have number of Y 
  &   Arrays
 & elements
 & Arrays can have any number of elements.

 \\\hline
 X combination of Y 
  &   green
 & blue
 & Green is a combination of blue and yellow.
.
 \\\hline
 X collection of Y
  &   society
 & individual
 & Society is now a collection of individuals

 \\\hline
 X branch of Y 
  &  Chinese medicine
 & Medical Qigong
 & Medical Qigong is a branch of traditional Chinese medicine

 \\\hline
 Y of X
  &   government
 & member
 & The prime minister shall inform all members of the government

  \\\hline
 Y ingredient in X
  &   Ephedrine
 & Ephedra
 & Ephedra is a key ingredient in Ephedrine

 \\\hline
 Y with X
  &   peacock
 & overgrown beak
 & I have an Indian Blue peacock with an overgrown beak

 \\\hline
 Y member of X 
  &   United Nations
 & Israel
 & Israel is a member of the United Nations.

 \\\hline
 X include Y 
  &   Symptom
 & vomiting
 & Symptoms can include vomiting
\\\bottomrule
 \end{tabular}}
\end{table}

\begin{table}[h!]
\footnotesize

\caption{All Hyponym-Hypernym patterns extracted by \name}\label{hypo-patterns-appendix}
\setlength{\tabcolsep}{1.5mm}
\small
{\begin{tabular}{p{0.18\linewidth}|p{0.15\linewidth}|p{0.15\linewidth}|p{0.45\linewidth}}\toprule
\bf Pattern & \bf X & \bf Y & \bf Sentence \\
\midrule
 X, a class of Y  &  Core 2 Duo & microprocessor & Core 2 Duo, a class of early Desktop micro-processor had much lower core frequency and approximately the same FSB frequency and level 2 cache size as Pentium D microprocessors \\\hline
 
 A class of Y, X
  &   Core 2 Duo & microprocessor &  A class of early Desktop micro-processor, Core 2 Duo, had much lower core frequency and approximately the same FSB frequency and level 2 cache size as Pentium D microprocessors. \\\hline
  
  X be a class of Y
  &   Core 2 Duo & microprocessor & Core 2 Duo is a class of early Desktop micro-processor which had much lower core frequency and approximately the same FSB frequency and level 2 cache size as Pentium D microprocessors.\\\hline
  
   X, a family of Y &{Vinyasa} &  {yoga} & {Vinyasa, a family of yoga is dynamic and}\\
   A family of Y, X &&&{ever-flowing.}\\
   X be a family of Y &&&\\ \hline
  X, a type of Y &  system  & computer  & The system software, a type of computer  \\
  A type of Y, X &software&software&software is designed for running the\\
  X be a type of Y &&&computer hardware parts and the application programs\\  \hline
  
  X, a kind of Y &  panda & bear & Panda, a kind of bear is found only in \\
  A kind of Y, X &&&China.\\
  X be a kind of Y &&&\\\hline
  
  Y, including X  &  Asiatic black & bear & Some species of bears, including Asiatic  \\
  Y which/that  &bear &&black bears and sun bears, are also\\
  include X&&&threatened by the illegal wildlife trade. \\
  Y include X &&&
  
  \\\hline
  
  Y, such as X &  sheep & domesticated  & Domesticated animals, such as sheep or  \\
  Y, for example X &&animal&rabbits, may have agricultural uses for\\
  Y, like X &&&meat, hides and wool.\\
  like many Y, X &&&\\\hline
  
  The Y of X
  &   Aasu & village & The village of Aasu along with Aoloau are jointly called O Leasina\\\hline
  
  X be Y &   panda & bear & The giant pandas are true bears, and part \\
   X be the Y&&&of the family Ursidae\\
   X be a Y &&&\\\hline
   
  X become Y
  &   kizzy
 & singer
 & In 2005, Kizzy became the lead singer of the "Bo Winiker Orchestra" with whom she performed for Bill Clinton, Glenn Close and with whom she gained critical acclaim for performing songs in Hebrew.
 \\\hline 
 Y named X \newline
 Y called X &  ponikve
 & village
 & Like other villages named Ponikve and similar names, it refers to a local landscape element.

 \\\hline 
Y as X
  &   Emperor
 & band
 & Since the 1990s, Norway's export of black metal, a lo-fi, dark and raw form of heavy metal, has been developed by such bands as Emperor, Darkthrone, Gorgoroth, Mayhem, Burzum and Immortal.

\\\hline 
Y "X"
  &   Clarens
 & village
 & A commission was appointed in 1912 to finalize negotiations, and a decision was made to name the village "Clarens" in honour of President Paul Kruger influence in the area.

\\\bottomrule
\end{tabular}}
\end{table}

\begin{table}[h!]
\caption{All Cause-Effect patterns extracted by \name}

\label{cause-effect-appendix}
\setlength{\tabcolsep}{1.5mm}
\footnotesize
{\begin{tabular}{p{0.19\linewidth}|p{0.15\linewidth}|p{0.15\linewidth}|p{0.4\linewidth}}\toprule
\bf Pattern & \bf X & \bf Y & \bf Sentence \\
\midrule
 Y caused by X &  sorafenib 
 & severe interstitial 
 & In this article, we describe a japanese patient with   
 \\
 X cause Y &treatment&pneumonia&severe interstitial pneumonia probably caused by\\
 X be a cause of Y  &&&sorafenib treatment for metastatic renal cell\\
 X be causes of Y &&&carcinoma.\\\hline

 Y induced by X  &   mizoribin 
 & A case of siadh 
 & A case of SIADH induced by mizoribin \\
 X induce Y &administration&&administration.
  \\\hline
  
  X lead to Y
  &   peripheral neuropathy
 & linezolid
 & However, peripheral neuropathy and bone marrow depression led to linezolid withdrawal in seven patients, and neuropathy may not be fully reversible in all patients.

 \\\hline
 Y be associated with X 
  &   sulfasalazine
 & pulmonary infiltrates
 & Pulmonary infiltrates and skin pigmentation are associated with sulfasalazine.
.
 
 \\\hline
 Y related to X 
  &   flecainide
 & interstitial hypoxaemiant pneumonitis
 & We describe a case of interstitial hypoxaemiant pneumonitis probably related to flecainide in a patient with the LEOPARD syndrome, a rare congenital disorder.
.\\\hline
 
X result in Y &   flucloxacillin
 & fatal hepatic injury
 & It is well-recognized that flucloxacillin may \\
Y be result of X&&&occasionally result in fatal hepatic injury. 
 \\\hline
 Y from X 
  &   exertion
 & satisfaction
 & I have always drawn satisfaction from exertion, straining my muscles to their limits.
.
\\\hline
 Y be triggered by X  &   earthquake
 & tsunami
 & A large tsunami is triggered by the earthquake \\
 X trigger Y &&&spread outward from off the Sumatran coast.
  \\\hline
 Y come from X 
  &   fear
 & blockage
 & Sometimes the blockage comes from fear, as for a CEO who hates public speaking but must give frequent speeches.
.
 \\\hline
 Y be the effect of X &   acupuncture
 & pain relief
 & Pain relief is the effect of acupuncture which lasts \\
 Y, the effect of X &&&for an extended period of time, sometimes months  \\
 &&&after the needle was removed.
  \\\hline
 X produce Y &   Ambient  
 & irritation
 & Ambient vanadium pentoxide dust produces  \\
 Y produced by X &vanadium
 pentoxide dust&& irritation of the eyes, nose and throat.\\\hline
 X promote Y 
  &   antiwar demonstrators
 & positive values
 & He created and advocated flower power,"a strategy in which antiwar demonstrators promoted positive values like peace and love to dramatize their opposition to the destruction and death caused by the war in Vietnam."
  \\\hline
 X generate Y  &   tunable laser
 & optical signal
 & The optical signal is generated by a tunable laser. \\
 Y generated by X &&&
 
 \\\hline
 X influence Y  &   tumorigenicity of clones
 & immunoprotective effects
 & The tumorigenicity of clones may be influenced by immunoprotective effects. \\
 Y influenced by X &&&
 
 \\\hline
 Y due to X &   Incorrect design
 & Failure in physical containment
 & Failures in physical containment may occur due to incorrect design. \\
 Y because of X &&&
  
\\\bottomrule
 \end{tabular}}
\end{table}

\end{document}